\documentclass{article}
\usepackage{arxiv}

\usepackage[utf8]{inputenc} 
\usepackage[T1]{fontenc}    
\usepackage{hyperref}       
\usepackage{url}            
\usepackage{booktabs}       
\usepackage{amsfonts}       
\usepackage{nicefrac}       
\usepackage{microtype}      
\usepackage{lipsum}		
\usepackage{graphicx}
\usepackage{natbib}
\usepackage{doi}
\usepackage[table]{xcolor}
\usepackage{amsmath}
\usepackage{amssymb}
\usepackage{multirow}
\usepackage[most]{tcolorbox}
\usepackage{xcolor}

\title{MedScribe: Clinically Grounded CT Reporting through Agentic Workflows}

\author{ \href{https://orcid.org/0009-0001-4188-2893}{\includegraphics[scale=0.06]{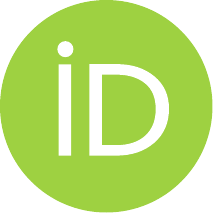}\hspace{1mm}Giuseppe A. Orlando} \\
	Inria Centre at Université Côte d'Azur\\
	Sophia Antipolis, France \\
	\texttt{giuseppe.orlando@inria.fr} \\
 \And
	\href{https://orcid.org/0000-0003-0651-4128}{\includegraphics[scale=0.06]{orcid.pdf}\hspace{1mm}Paolo Papotti} \\
	EURECOM\\
	Biot, France \\
  \And
	\href{https://orcid.org/0000-0002-1147-766X}{\includegraphics[scale=0.06]{orcid.pdf}\hspace{1mm}Maria A. Zuluaga} \\
	EURECOM\\
	Biot, France \\
 \And
	\href{https://orcid.org/0000-0003-4440-5416}{\includegraphics[scale=0.06]{orcid.pdf}\hspace{1mm}Olivier Humbert} \\
	Dep. of Nuclear Medicine, Centre Antoine Lacassagne\\ Nice, France\\
    Université Côte d’Azur, CNRS, Inserm, iBV \\
	Nice, France \\
	\And
	\href{https://orcid.org/0000-0003-0521-2881}{\includegraphics[scale=0.06]{orcid.pdf}\hspace{1mm}Marco Lorenzi} \\
	Inria Centre at Université Côte d'Azur\\
	Sophia Antipolis, France \\
}

\date{}

\renewcommand{\headeright}{}
\renewcommand{\undertitle}{}
\renewcommand{\shorttitle}{MedScribe}

\hypersetup{
pdftitle={MedScribe: Clinically Grounded CT Reporting through Agentic Workflows},
pdfsubject={},
pdfauthor={Giuseppe ~Orlando, Paolo ~Papotti, Maria ~Zuluaga, Olivier ~Humbert, Marco ~Lorenzi}
pdfkeywords={CT Report Generation, Agentic LLMs, Segmentation},
}

\begin{document}
\maketitle

\begin{abstract}
Vision-language models (VLMs) have shown potential for automated radiology report generation, yet existing approaches rely on global embedding compression of volumetric data, often leading to hallucinated findings and limited anatomical grounding in 3D CT imaging. We introduce MedScribe, a hypothesis-driven framework that reformulates report generation as an iterative evidence acquisition process rather than a single-pass encoding task.
MedScribe models reporting as a sequential decision process in which a large language model dynamically invokes pathology-specific diagnostic tools to extract localized volumetric features. These structured features are used to query a multidimensional retrieval space aligned with pathology-specific textual evidence. By explicitly accumulating quantitative evidence prior to synthesis, the framework enforces fine-grained grounding and reduces unsupported claims.
Without task-specific fine-tuning, MedScribe improves clinical accuracy, factual consistency, and interpretability on CT-RATE and RadChestCT compared to state-of-the-art 2D and 3D VLMs, demonstrating the value of hypothesis-driven reasoning for reliable medical image reporting.
\end{abstract}


\section{Introduction}
\label{sec:intro}

Medical Vision-Language Models (VLMs) are attracting increasing attention for their potential to assist clinicians in radiology reporting by generating structured textual descriptions directly from imaging data. By combining large-scale visual encoders with autoregressive language decoders, these systems aim to translate image content into clinically meaningful narratives. While recent advances have significantly improved linguistic fluency, current medical VLMs remain limited in their ability to produce factually grounded and clinically reliable reports. In particular, generated descriptions often lack precise correspondence with the underlying pathological evidence present in the image, raising concerns about hallucinated findings and unreliable reasoning.

The domain of radiology report generation initially relied on 2D architectures, especially thanks to the availability of large-scale datasets such as MIMIC-CXR \citep{johnson2019mimic}. Early frameworks approached the task by mapping static image features to sequential text, allowing for the production of foundation models like R2Gen \citep{chen2020generating}, BioViL \citep{boecking2022making}, BioViL-T \citep{bannur2023learning}, and MedCLIP \citep{wang2022medclip}. 
While these models demonstrated promising cross-modal alignment capabilities, they remained confined to the 2D imaging paradigm, falling short of capturing the volumetric context essential for interpreting complex 3D modalities such as computed tomography (CT) scans. Extending VLMs to volumetric CT data introduces substantial challenges. Standard ViT-based encoders require aggressive downsampling or slice sampling strategies to accommodate memory and context constraints, creating an information bottleneck that can corrupt subtle pathological patterns. Furthermore, radiology reports provide sparse supervision relative to the full 3D volume, as they typically describe only salient abnormalities. These factors complicate the learning of fine-grained anatomical reasoning and robust image-text grounding.

The recent release of the CT-RATE dataset \citep{hamamci2024developing} marked a significant milestone, providing a large repository of paired volumetric chest CT images, reports, and pathology labels. This paved the way to the development of both generalist models, such as CT-CHAT \citep{hamamci2024developing}, and report generation models such as CT2Rep \citep{hamamci2024ct2rep}. While these models rely on sophisticated multimodal fusion and pre-training strategies to link 3D representations with textual descriptions, they still inherit the information bottleneck of ViT-based encoders, and critically, cannot explicitly verify which anatomical regions drive a generated finding.

Recently, models such as VILA-M3 \citep{Nath_2025_CVPR} have attempted to enhance medical VLMs by incorporating specialized medical expert knowledge into the model's training phase. Simultaneously, generalist models like Gemini offer native multimodality and massive context windows for complex reasoning \citep{comanici2025gemini25pushingfrontier}, while Med-Gemma provides accessible clinical grounding \citep{sellergren2026medgemma15technicalreport}. However, these models rely on passive visual encoding and lack mechanisms to verify clinical hypotheses against volumetric data. Furthermore, as we demonstrate in this work, they lack the anatomical depth required for fine-grained pathological descriptions.  

In this work, we propose to reformulate 3D radiology report generation as an iterative reasoning and evidence-querying process rather than a single-pass encoding task. In particular, we introduce hypothesis-driven volumetric evidence acquisition as a new paradigm for 3D report generation.

Inspired by the Reasoning and Acting (ReAct) framework \citep{yao2023react} we introduce MedScribe, a tool-augmented 3D VLM that integrates structured reasoning with targeted volumetric evidence extraction. Instead of compressing the entire CT scan into a fixed embedding, MedScribe employs the language model as a central reasoning agent that dynamically invokes specialized tools to retrieve quantitative and spatial descriptors, including volumetric measurements, intensity statistics, and anatomical coordinates.

By querying specific visual evidence in a targeted, iterative manner, MedScribe aligns clinical reasoning with the volumetric data, ensuring factual grounding that is currently not achievable with static end-to-end architectures, while not requiring any task-specific fine-tuning.

Overall, our contributions are: (i) a novel agentic framework for 3D report generation that decouples linguistic reasoning from visual perception through iterative tool invocation; (ii) a multidimensional RAG space grounded in quantitative radiomics descriptors; and (iii) two operational modes demonstrating framework generality. Evaluations on CT-RATE and RadChestCT show superior clinical accuracy and factual grounding over state-of-the-art VLMs.

\section{Methods}
\label{sec:methods}
\begin{figure*}[t]
  \centering
  \includegraphics[width=0.9\textwidth]{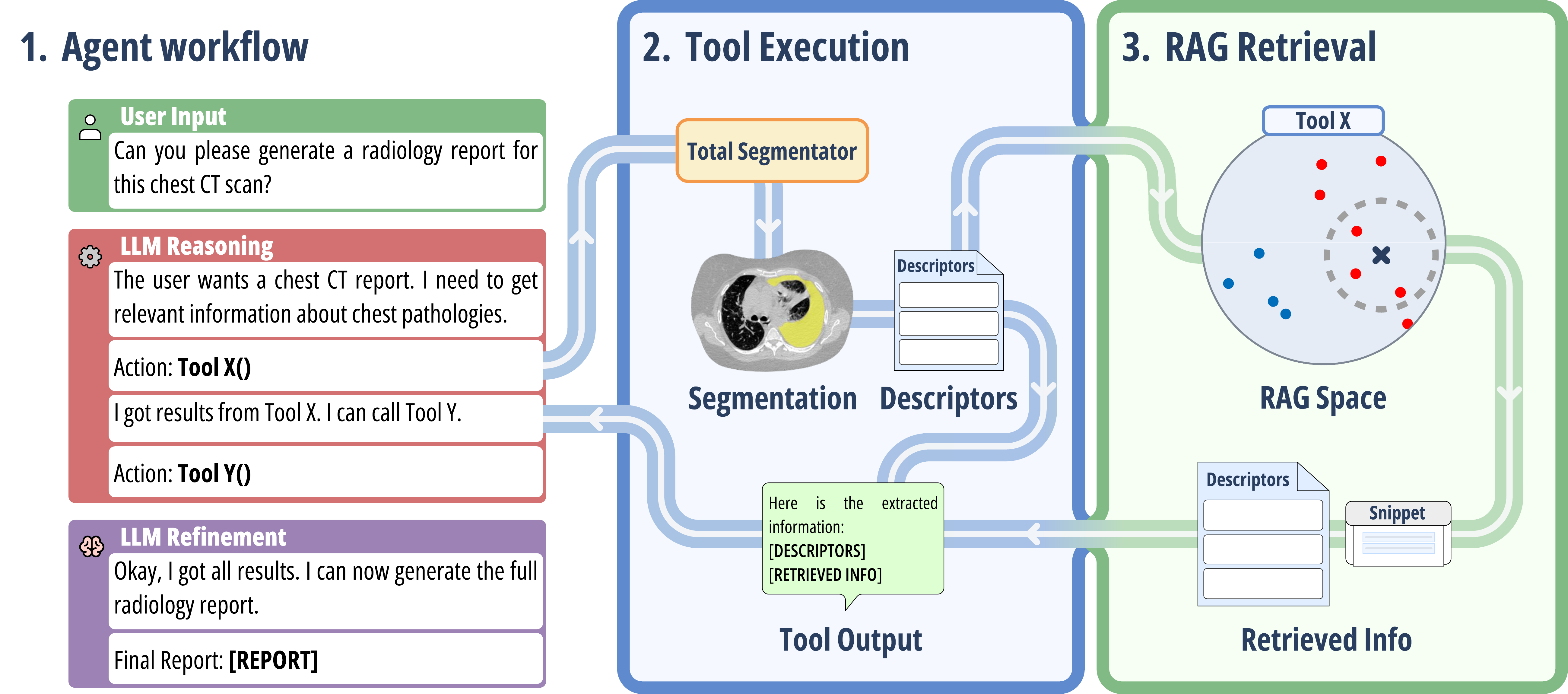}
  \caption{Overview of MedScribe.  Agent Workflow: Triggered by a user query, the LLM first goes to a Reasoning Phase to obtain pathology information by calling available tools and then ends in the Refinement Phase where it builds the final response.}
  \label{fig:pipeline}
\end{figure*}

Let $\mathbf{V} \in \mathbb{R}^{H \times W \times D}$ denote a CT volume and $q$ a reporting query. 
Standard VLMs generate reports via a single-pass mapping, $r = g(h(\mathbf{V}), q)$, where $h(\cdot)$ compresses the full 3D volume into a fixed embedding. Such global compression limits anatomical grounding and prevents explicit verification of intermediate clinical hypotheses. We instead formulate report generation as a \textit{sequential evidence acquisition process} (Figure \ref{fig:pipeline}), where the final report is generated after accumulating structured volumetric evidence:
\begin{equation}
r = g\big(q, \mathcal{E}_T\big),
\end{equation}
where $\mathcal{E}_T$ denotes the evidence collected over $T$ reasoning steps.

\subsection{Multi-Stage Agent Workflow}
\label{ssec:agentic_workflow}

MedScribe follows ReAct principles of thought and action based on a standard LLM that has access to a set of imaging tools capable of extracting specific features derived from anatomical segmentation and radiomics analysis. 

At step $t$, the agent state is defined as $s_t = (q, \mathcal{E}_t)$, with $\mathcal{E}_0 = \varnothing$. 
The agent selects actions $a_t \in \mathcal{A} \cup \{\texttt{STOP}\}$ with $a_t \sim \pi(a \mid s_t)$, 
where $\mathcal{A}$ is a set of pathology-specific diagnostic tools, and $\pi$ is  implemented via prompted LLM reasoning constrained to select from the tool set 
$\mathcal{A}$ (Section \ref{ssec:tool_invocation}). If $a_t \neq \texttt{STOP}$, a localized feature extractor is applied: $\mathbf{f}_t = \phi(\mathbf{V}, a_t)$, producing structured volumetric descriptors $\mathbf{f}_t \in \mathbb{R}^d$ derived from segmentation and radiomics analysis. In particular, $\mathbf{V}$  is only accessed locally through the tool calls.
The structured feature vectors  $\mathbf{f}_t$ are then used to query a multidimensional RAG retrieval space on the CT-RATE \textit{reference dataset} (Section \ref{ssec:rag_space}). This last step allows the LLM to receive a grounded visual-to-textual mapping consisting of both tool-derived quantitative features and retrieved reference samples, and to update the evidence $\mathcal{E}_{t+1}$  for the next reasoning step.


When the reasoning loop is concluded, MedScribe generates the final response $r = g(q, \mathcal{E}_T)$ through a synthesis phase. We formulate two distinct operational settings. In the first one, \textit{template-based augmentation}, the agent begins with a neutral boilerplate report to initialize an uninformative evidence set $\mathcal{E}_0$ (MedScribe-Template). In the second setting, \textit{VLM refinement}, the agent acts as an automated reviewer: it first queries a baseline VLM to initialize the evidence set $\mathcal{E}_0$, and subsequently invokes the pathology tools to verify and eventually correct those findings (MedScribe VLM).

\subsection{Diagnostic Tool Invocation Pipeline}
\label{ssec:tool_invocation}
To bridge the gap between the textual reasoning of the LLM and the visual complexity of the CT scans, MedScribe relies on a structured tool calling mechanism. 
Depending on the state $s_t$, the agent determines the tools associated with action $a_t$ to produce descriptors $\mathbf{f}_t = \phi(\mathbf{V}, a_t)$. This quantitative representation serves a dual purpose: first, it grounds the LLM's diagnostic reasoning to strict numerical data; second, it acts as the query vector required for our RAG reference space (Section~\ref{ssec:rag_space}). The toolset $\mathcal{A}$ triggered by the agent is associated with 86 dimensions related to anatomical and pathological segmentation of structures commonly analyzed in chest CT imaging like lungs, heart, arteries and effusions (obtained via validated models for multi-organ chest CT segmentation, e.g. TotalSegmentator \cite{wasserthal2023totalsegmentator}). We also provide imaging features from these masks combined with the original images of the reference dataset, including absolute and relative volumes (calculated relative to reference organs like the lungs and heart), as well as axial extent and thickness. To capture spatial distribution of pathologies like effusions, masks were partitioned along anatomical midlines to derive laterality (left/right) and orientation (anterior/posterior) fractions. For findings requiring intensity analysis, such as calcification and nodules, Hounsfield Unit (HU) values were extracted from the bounded regions.

\subsection{Multidimensional RAG retrieval space}
\label{ssec:rag_space}
Unlike conventional RAG systems that retrieve based on text embeddings or learned visual embeddings \cite{RadAlign,pmlr-v219-ranjit23a}, our retrieval space is conditioned on quantitative anatomical descriptors and  textual information from paired images and reports of our \textit{reference dataset}.

To ground reasoning, we retrieve similar reference cases from a database of paired features and text $(\mathbf{f}_j^{ref},\mathbf{t}_j^{ref})_{j=1}^M$ from the reference database, where $\mathbf{t}_j^{ref}$ is the pathology-specific textual snippet obtained via LLM four-shot prompting. For each of the 18 target pathologies, the LLM is instructed to isolate exact sentences from the Findings sections, ensuring that $\mathbf{t}_j^{ref}$ describes single pathologies. All the negative cases, if no description of a given pathology is found in the report, are mapped to a simple \textit{"no finding"} sentence. This extraction was human curated to ensure high fidelity and is empirically verified against ground-truth labels. The full prompt template, an illustrative set of examples, and the empirical verification of the extracted snippets are provided in Supplementary Section~\ref{supp:prompt}. For an input feature vector $\mathbf{f}_t$, the retrieval is performed via nearest-neighbor search in the structured feature space 

$$\mathcal{N}_k(\mathbf{f}_t) = \{\mathbf{t}_j^{ref} \mid j\in I_k(\mathbf{f}_t)\},$$ 

where $I_k(\mathbf{f}_t)$ is the set of top $k$ closest feature vectors of the reference dataset, $I_k(\mathbf{f}_t) = \operatorname*{arg\,topk}_{j \in \{1,\dots,M\}}
 \left(
 - d(\mathbf{f}_j^{ref}, \mathbf{f}_t)
 \right)$, for a given distance $d$. As a design choice, we consider here a standard $L^2$ distance on the feature vectors of standardized features. The evidence set is then updated as $\mathcal{E}_{t+1} = \mathcal{E}_t \cup \{\mathbf{f}_t\} \cup \mathcal{N}_k(\mathbf{f}_t)$, and reasoning terminates with $a_t = \texttt{STOP}$ when the LLM determines that sufficient evidence has been accumulated, either after querying all relevant pathologies, or earlier if observations indicate a finding is absent or uninformative.

\section{Experimental Setup}
\label{sec:setup}

To evaluate the efficacy of our expert guided ReAct framework we conducted experiments across two large-scale 3D CT datasets, comparing with different state-of-the-art VLMs, performing ablation studies on each component of MedScribe, and by developing a new benchmark for fine-grained anatomical assessment.

\medskip
\noindent\textbf{Datasets.}
Our benchmark is based on two primary 3D chest CT repositories. CT-RATE, the largest publicly available dataset of paired 3D chest CT scans and radiology reports, serves as our primary development and evaluation source. Specifically, we use the CT-RATE training split ($\sim$20,000 scans) to populate our multidimensional RAG reference space, providing a diverse and expansive search space for feature-to-text mapping. The CT-RATE validation set (1,304 scans) is used as our primary test set for report generation performance.
To assess the generalizability of our framework, we further evaluate performance on the test set of RadChestCT (360 samples). To ensure consistency across datasets, we performed a label mapping process to align the abnormalities in RadChestCT with the pathology definitions used in CT-RATE, allowing for a direct comparison of the expert guided reasoning loop.

\medskip
\noindent\textbf{Implementation and Framework Configuration.}
Our framework is built using the LangChain library \cite{langchain}, which provides the orchestration layer for our ReAct based agent. We utilize Llama 3.1-8B \cite{grattafiori2024llama3herdmodels} in three distinct capacities: as the extractor to identify pathology findings from reports during the preparation phase in Section \ref{ssec:rag_space}, as the central orchestrator to decide which expert tools are suitable for the user input, and as the refiner to later understand tool output and compose the response to the user input. The choice of Llama 3.1-8B is motivated by the need to match the LLM backbone of CT-CHAT \cite{hamamci2024developing}, isolating the contribution of the agentic tool-grounded design from the underlying LLM strength. To verify that the conclusions are not specific to this scale, we additionally tested MedScribe-Template with the larger Llama 3.3-70B backbone (Supplementary Section~\ref{supp:llm_backbone}), confirming that the framework benefits monotonically from stronger LLMs (+9.29\% in MacroF1 score averaged over all pathologies) while remaining effective at the 8B scale. In all three capacities, the LLM is queried with temperature set to zero to ensure deterministic and reproducible behaviour across runs.

The visual extraction pipeline in Section \ref{ssec:tool_invocation} is powered by TotalSegmentator v2 \cite{wasserthal2023totalsegmentator}, which generates segmentation masks for different target structures including the heart, lungs, arteries, lung nodules and effusions. From these regions, we derive a comprehensive feature set consisting of volumetric measurements, density analysis (mean and percentile Hounsfield Units), and spatial coordinates.

The RAG component utilizes a k-Nearest Neighbors (k-NN) approach to retrieve the top $k=3$ reference samples based on feature space proximity. This value was empirically selected to provide the LLM with sufficient linguistic variety to understand how to translate quantitative features into text while keeping the amount of in-context information low. To confirm the robustness of this choice, we also performed a sensitivity analysis across $k \in \{1, 3, 5, 7, 9, 11, 13\}$ on a smaller set of the CT-RATE validation set (Supplementary Section~\ref{supp:k_ablation}), observing that classification and text-generation performance remain stable within a narrow band across the tested range. All experiments were conducted on a single NVIDIA H100 GPU to allow inference with the larger baseline comparison models.

We compared our framework against several SOTA models, including Gemini 2.5 Flash \cite{comanici2025gemini25pushingfrontier}, MedGemma \cite{sellergren2026medgemma15technicalreport}, VILA-M3 \cite{Nath_2025_CVPR} and CT-CHAT \cite{hamamci2024developing}. To quantify the general understanding capabilities of the models we focused primarily on the task of full radiology report generation. Moreover, to assess the understanding of fine-grained pathological details, we introduced an additional benchmark targeting the accuracy of laterality prediction (left/right) for pleural effusion pathology. This assessment is motivated by the localized nature of this pathology and by the fact that laterality is explicitly reported in the ground-truth radiology reports for 302 scans in the CT-RATE validation set, which form the evaluation subset for this task. 
Since these annotations are associated with cases that show clear signs of effusion on either sides, the reported performance should be interpreted as an indicator of the model's behaviour on clear cases, and not of the pathology in general.


\medskip
\noindent\textbf{Baseline configuration and fairness.}
All baselines were queried under the best configuration we could identify through preliminary experiments, with input preprocessing matched as closely as possible across models to ensure a fair comparison. Both MedGemma and Gemini natively support multi-image inputs for volumetric medical imaging: MedGemma 1.5-4B-it is explicitly designed for three-dimensional CT and MRI volume interpretation via sequences of axial slices \cite{sellergren2026medgemma15technicalreport}, while Gemini 2.5 Flash supports up to 3,600 image inputs per request, comfortably accommodating volumetric CT as a slice sequence. For both models, we performed a slice-count ablation across $n_{\text{slices}} \in \{16, 32, 64, 128\}$ and selected the best-performing configuration ($n_{\text{slices}}=128$ for both, see Supplementary Section \ref{supp:baselines}). Higher number of slices configurations were not considered because 1) they were exceeding the recommended amount of slices for MedGemma inference ($\sim$85 slices) \cite{sellergren2026medgemma15technicalreport}  while also exceeding the memory constraints of our hardware setup. 
Image preprocessing followed MedGemma's official CT windowing specification, which we then applied consistently to Gemini to ensure both 2D-based models observed identical inputs. CT-CHAT, the only fully 3D-native baseline, was queried using the exact prompt and preprocessing pipeline specified in its public implementation. MedScribe, MedGemma, and Gemini were queried with the same instruction prompt targeting full report generation.

To evaluate the generated reports, pathology labels were derived using a finetuned RadBERT \cite{yan2022radbert} classifier, the same used in the creation of the CT-RATE dataset; performance of all baselines is then evaluated in terms of F1 classification score. On top of the F1 scores we also provide some classical text metrics to compare the fluency of the different baselines: BLEU, ROUGE, METEOR and F1-RadGraph \cite{delbrouck-etal-2024-radgraph}. The full source code used in this study will be made publicly available upon acceptance of the manuscript.

\section{Results}
\label{sec:results}

\begin{figure*}[t]
  \centering
  \includegraphics[width=\textwidth]{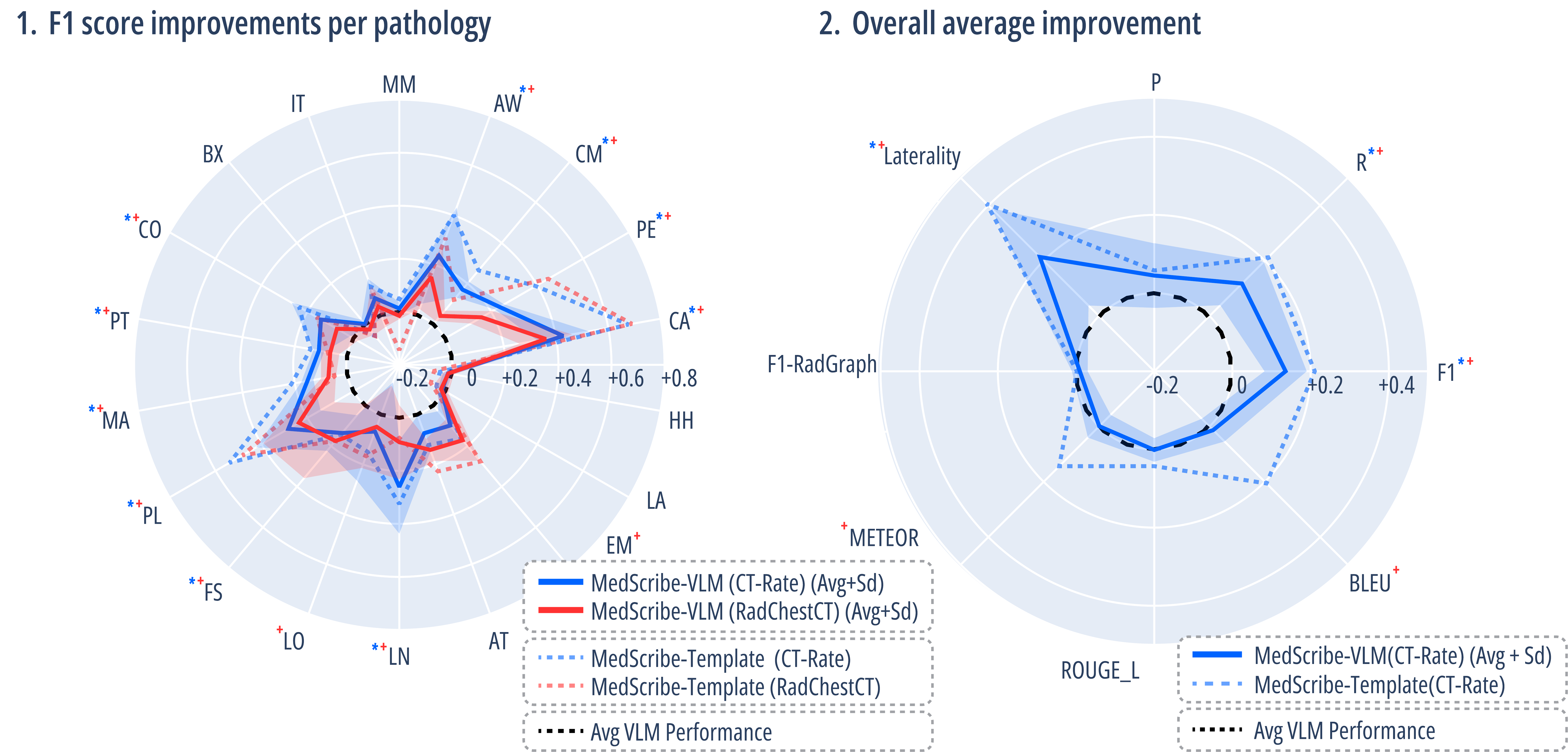}
  \caption{Comparative analysis of MedScribe configurations against baseline VLMs. (Left) Per pathology classification F1 score improvements of both  \textit{template-based augmentation} (MedScribe-Template) and  \textit{VLM refinement} (MedScribe-VLM) settings  across the 18 CT-RATE pathologies \cite{hamamci2024developing}. (Right) Average pathology classification and text metrics improvements of both our configurations on the CT-RATE dataset. Stars ($\textcolor{blue}{*}$) and plus signs ($\textcolor{red}{+}$) denote statistically significant improvements ($p<0.05$, linear mixed-effects model; see Supplementary Section~\ref{supp:stat_analysis}) for MedScribe-Template and MedScribe-VLM over baseline VLMs, respectively.}
  \label{fig:performance}
\end{figure*}

\textbf{Report generation improvement.} Our findings are summarized in Figure \ref{fig:performance}, illustrating per-pathology F1 scores  (left panel) and the overall metrics (right panel) when comparing head-to-head MedScribe-VLM to each of the 4 baseline VLM, and MedScribe-Template to the average VLM performance.  
Figure \ref{fig:performance}, left panel, shows the average absolute improvement of MedScribe in F1 scores across all evaluated pathological classes when compared to the averaged VLM baselines. 
When MedScribe is deployed in both \textit{VLM refinement} (MedScribe-VLM, continuous lines) and \textit{template-based augmentation}  (MedScribe-Template, dashed lines) we observe a consistent  increase in performance across nearly all pathologies compared to the standalone baseline, statistically significant for the majority of the   pathologies ($p<0.05$). The specific cases where MedScribe does not show a clear performance gain correspond to pathologies that are not correctly assessed by the diagnostic tool invocation pipeline (Section \ref{ssec:tool_invocation}), suggesting that the performance gains are driven by the grounding provided by the targeted feature extraction and subsequent RAG retrieval space rather than LLM hallucinations.

\begin{figure*}[t]
  \centering
  \includegraphics[width=\textwidth]{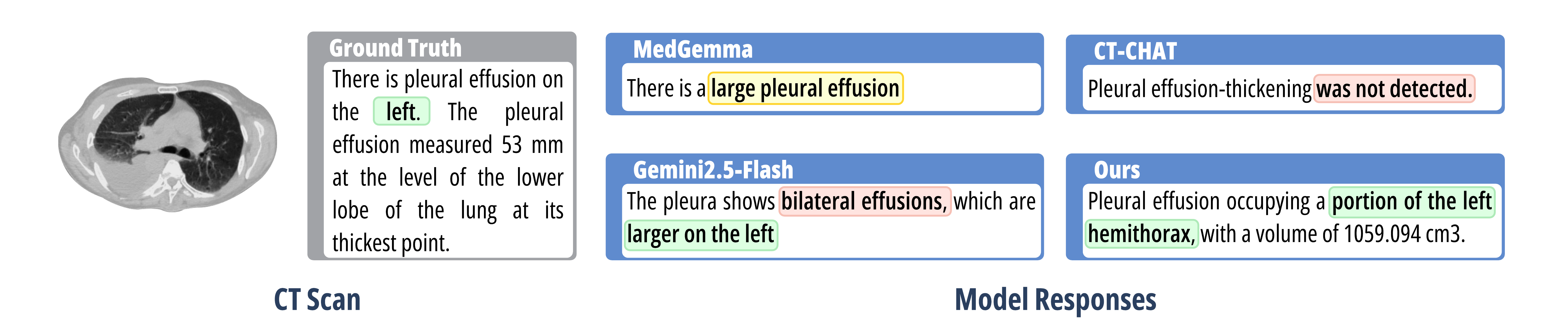}
  \caption{Fine-grained understanding of pleural effusion laterality in report generation. Correct and incorrect findings are highlighted respectively in green and red.}
  \label{fig:laterality}
\end{figure*}

The right panel of Figure \ref{fig:performance} compares MedScribe's performance against the baselines by plotting average classification and text metrics on the CT-RATE dataset. We note that the MedScribe-Template setting consistently outperforms all other baselines across metrics. In particular, baseline VLMs perform noticeably worse on the laterality task, indicating improved fine-grained understanding of MedScribe (qualitative assessment in Figure \ref{fig:laterality}). The complete numerical results behind Figure~\ref{fig:performance}, including per-pathology F1 scores for all baseline VLMs, all MedScribe-VLM configurations, and MedScribe-Template on both CT-RATE and RadChestCT, together with the legend of the pathology abbreviations used in the supplementary tables, are reported in Supplementary Section~\ref{supp:detailed_results}.


\begin{table}[b]
\caption{Ablation study: Average classification and text scores on CT-RATE. All the configurations are run in the template based setting.}
\centering
\resizebox{\textwidth}{!}{%
\begin{tabular}{l *{8}{c}}
\midrule
\rowcolor{gray!10}
\multicolumn{9}{c}{\textbf{Dataset: CT-RATE}} \\
\midrule
& Precision & Recall & F1-score & Bleu & Rouge & Meteor & RadGraphF1 & LateralityF1 \\ 
\midrule
Ours & 0.36 & 0.33 & \textbf{0.32} & \textbf{0.34} & \textbf{0.18} & \textbf{0.34} & \textbf{0.12} & 0.59 \\
Ours (NO RAG) & 0.18 & 0.60 & 0.24 & 0.21 & 0.12 & 0.28 & 0.08 & 0.56 \\
Ours (NO React) & 0.28 & 0.33 & 0.26 & 0.31 & 0.17 & 0.29 & 0.11 & \textbf{0.70} \\
Ours (Tools + Dummy report) & 0.34 & 0.32 & 0.27 & 0.01 & 0.05 & 0.07 & 0.05 & Not supported\\
Ours (XGBoost + Dummy report) & \textbf{0.51} & \textbf{0.43} & 0.29 & 0.02 & 0.04 & 0.07 & 0.05 & Not supported\\
\midrule
\end{tabular}
}
\label{tab:ablation}
\end{table}

\noindent

\textbf{Ablation results.} Table \ref{tab:ablation} reports the results for MedScribe ablation study. We first ablated the RAG retrieval space, forcing the agent to rely on raw extracted features. This caused an overall drop in accuracy showing the crucial role of reference sample retrieval (Recall=0.60, Precision=0.18). At $0.18$ precision, the majority of predicted positives are false, which is clinically undesirable as it undermines radiologist trust and inflates downstream review burden. The F1 score, which jointly accounts for missed and spurious findings, is therefore a more clinically appropriate headline metric than recall in isolation. We also ran our methods in a non-agentic way, providing all tools' results together with the user query. This configuration is associated with generally lower classification and text generation performances, indicating lower reliability in the report generation task. Interestingly, the laterality test F1 score is slightly higher, likely due to the fact that laterality is a spatially localized single-step judgment that benefits from having complete bilateral evidence.

We finally tested two \textit{dummy report} baselines that replace the LLM synthesis with rule-based text sequences, using either majority voting or an XGBoost classifier to predict present pathologies from the localized features. While these two configurations are comparable to MedScribe in classification, they fail on text generation and laterality tasks, demonstrating the need for the LLM to generate high-quality final response. 

We note that the reported macro F1 (0.32) should be interpreted in light of the 18-pathology averaging and the heterogeneity of the target classes: for findings with dedicated tools and sufficient training support, MedScribe reaches substantially higher per-pathology F1 scores, while pathologies without tool coverage (e.g., hiatal hernia, lymphadenopathy) pull the macro average down. Per-pathology performance is reported in Figure~\ref{fig:performance} and Supplementary Table~\ref{tab:per_pathology_f1}. Regarding RadGraphF1, we note that all evaluated models score in a narrow low range (0.08--0.22), which is consistent with the metric being originally designed and validated on 2D chest X-ray reports \cite{delbrouck-etal-2024-radgraph}; its transferability to the longer, more heterogeneous 3D CT report setting remains an open methodological question for the community, rather than a MedScribe-specific limitation.

\medskip
\noindent\textbf{Sensitivity to segmentation errors.}
Since MedScribe's tools rely on segmentation masks to extract anatomical descriptors, the quality of the segmentation directly influences the reliability of the downstream retrieval and reasoning steps. To quantify this dependency, we identified scans for which lung segmentation is likely unreliable by flagging the top 10\% of the test set ($n = 259$) with abnormally large lung volumes; this is a strong indicator of oversegmentation, where the predicted mask extends beyond true lung boundaries and corrupts the downstream radiomics. On these samples, the radiomics-based neighbor search retrieves less relevant training examples, introducing noise directly into the retrieval step.

Table~\ref{tab:supp_seg_sensitivity} reports the resulting drop in report quality. Macro F1 falls from 0.323 on the full test set to 0.234 on the oversegmented subset ($-$8.9 points) when averaged across all 18 pathologies, and from 0.312 to 0.230 ($-$8.1 points) when evaluation is restricted to lung-specific pathologies (emphysema, atelectasis, consolidation, lung opacity, lung nodule, pulmonary fibrotic sequela, mosaic attenuation pattern, peribronchial thickening, bronchiectasis, interlobular septal thickening). The fact that the degradation is comparable across the two settings confirms that the error propagates specifically through the affected anatomical region: when the lung mask is unreliable, lung-specific radiomics are corrupted, retrieval returns less similar neighbors, and the final report quality degrades accordingly.

\begin{table}[h]
\caption{Sensitivity to segmentation errors. Macro F1 scores on the full CT-RATE test set compared to the subset of scans flagged as likely oversegmented (top 10\% by lung volume, $n=259$). Results are reported both across all 18 pathologies and restricted to lung-specific pathologies.}
\centering
\begin{tabular}{l c c c}
\midrule
\rowcolor{gray!10}
Evaluation & Full test set & Oversegmented subset & $\Delta$ \\
\midrule
All pathologies (18)            & 0.323 & 0.234 & $-$0.089 \\
Lung-specific pathologies (10)  & 0.312 & 0.230 & $-$0.082 \\
\midrule
\end{tabular}
\label{tab:supp_seg_sensitivity}
\end{table}

\section{Conclusion and Discussion}
\label{sec:discussion}

We introduced MedScribe, a novel agentic framework for 3D radiology report generation that decouples linguistic reasoning from visual perception through iterative tool invocation. 
The methodological innovation of MedScribe consists in making the RAG space a structured representation model, where retrieval is pathology-conditioned, and the feature space is anatomically factorized. This contribution stands apart from typical retrieval-augmented (RAG) systems based on the injection of static context, and from pipeline-based models like ToolFormer \cite{schick2023toolformer} that dispatch actions in a rigid feed-forward manner. Overall, MedScribe provides a natural, clinically motivated inductive bias for structured reporting while ensuring a transparent chain of evidence. Unlike recent agentic frameworks such as Ophiuchus \cite{jiang2025incentivizing} or MedAgent-Pro \cite{wang2025medagent}, which remain largely confined to 2D inputs or require extensive domain-specific adaptation, MedScribe leverages the zero-shot reasoning of frontier LLMs to achieve state-of-the-art results for complex 3D CT volumes. 

Regarding the results in the MedScribe ablation study (Table $\ref{tab:ablation}$), a classifier trained on the raw features achieves precision and recall higher than MedScribe (XGBoost + Dummy report). However, these results are susceptible to potential label circularity. RadBERT was used both to generate the labels for the CT-RATE dataset and to evaluate the resulting reports. MedScribe avoids this circularity reasoning over raw quantitative tool outputs and retrieving samples via feature space distances. Furthermore, the generalization results on RadChestCT provide a crucial validation: those labels were derived through an independent extraction pipeline rather than the RadBERT-based model used for CT-RATE. The fact that MedScribe’s performance gains persist under this different labels source proves that our improvements are not simply due to the possible model bias.

We note that the quality of the extracted features is directly conditioned on segmentation accuracy. We quantify this dependency in Section~\ref{sec:results}, showing that scans with unreliable lung segmentation incur a consistent and anatomically localized drop in report quality. This analysis has two implications. First, segmentation quality is a meaningful bottleneck of the current pipeline, and improving segmentor robustness is a direct path to improving report generation. Second, the observation that errors propagate through the specific anatomical region affected (rather than uniformly across pathologies) confirms the interpretability of the pipeline: the path from tool output to retrieved evidence to generated text is traceable, and failure modes are localizable to specific components rather than opaque.

Tool design and performance are also extremely coupled. MedScribe's toolset spans a spectrum of feature specificity: for pathologies with targeted imaging signatures (e.g., pleural and pericardial effusion, coronary and arterial wall calcification, cardiomegaly), dedicated tools produce pathology-tailored descriptors that drive sharp retrieval in the RAG space and yield the strongest gains. For pathologies without a dedicated extractor but with a characteristic anatomical substrate (e.g., lung opacity, emphysema, atelectasis, consolidation, lung nodules, bronchiectasis), general-purpose lung-region features still support meaningful retrieval, leading to more moderate but consistent improvements. The cases where MedScribe does not improve correspond to pathologies for which no current feature, specific or general, provides a meaningful imaging signature (e.g., hiatal hernia, lymphadenopathy). These are not failures of the framework itself but a direct consequence of its modular design since pathology coverage is a function of feature coverage. Our framework is extensible to broader tool libraries to capture currently under-represented findings, such as interstitial lung disease, mediastinal masses, or airway disease.

A further open aspect concerns the handling of contradictions and borderline findings. In the MedScribe-VLM configuration, the final synthesis step may encounter conflicts between the baseline VLM's initial draft and the tool-derived evidence; similarly, in both operational modes, tool outputs may lie near decision boundaries and the retrieved reference cases may disagree (e.g., two of the three neighbors indicate the presence of a finding and one its absence). The current framework does not explicitly handle these situations: the arbitration is delegated to the final LLM, which weighs the available evidence implicitly. A promising direction is to exploit the multidimensional RAG space itself as a source of uncertainty: the agreement among the $k$ retrieved neighbors for a given pathology provides a natural, interpretable confidence signal, which could be given to the LLM via the prompt and included in the generated report as an explicit uncertainty qualifier. Introducing such calibrated uncertainty communication represents a direct extension of the current framework, and would allow MedScribe to provide clinically actionable information.


\section*{Supplementary Material}
\label{sec:supplementary}

\setcounter{subsection}{0}
\renewcommand{\thesubsection}{S\arabic{subsection}}

\subsection{Detailed per-pathology and aggregated results}
\label{supp:detailed_results}

This section reports the complete numerical results that complement the spider-plot visualizations of Figure~\ref{fig:performance} in the main text. Table~\ref{tab:per_pathology_f1} provides the per-pathology F1 scores on both CT-RATE and RadChestCT for the four general-purpose VLM baselines (CT-CHAT, Gemini 2.5 Flash, MedGemma 1.5-4B-it, VILA-M3) in their standalone configuration and after MedScribe-VLM refinement, together with the standalone MedScribe-Template results. Tables~\ref{tab:summary_metrics_ctrate} and \ref{tab:summary_metrics_radchest} report the corresponding aggregated classification and text-generation metrics for CT-RATE and RadChestCT, respectively. The legend of the pathology abbreviations used in the column headers of Table~\ref{tab:per_pathology_f1} is reported below the table itself.

Two consistent patterns emerge from these tables. First, the addition of MedScribe to a baseline VLM (\textit{i.e.}, MedScribe-VLM mode) systematically improves classification performance over the standalone VLM, on both datasets and for nearly all pathologies, confirming that the tool-grounded evidence and the retrieved RAG snippets provide useful additional signal regardless of the underlying VLM. Second, the standalone MedScribe-Template configuration achieves the strongest macro F1 on both datasets, despite using no VLM at all and operating purely on the tool outputs and the retrieved textual snippets, which supports the view that the gains observed in MedScribe-VLM are primarily driven by the agentic, tool-grounded design rather than by the underlying VLM.


\begin{table*}[t]
\centering
\caption{Per-pathology F1 scores across models and datasets. Best result per column shown in \textbf{bold}.}
\label{tab:per_pathology_f1}
\resizebox{\textwidth}{!}{%
\begin{tabular}{lcccccccccccccccccc}
\midrule
\rowcolor{gray!10}
\multicolumn{19}{c}{\textbf{\textit{CT-RATE}}} \\
\midrule
\rowcolor{gray!10}
Model & MM & AW & CM & PE & CA & HH & LA & EM & AT & LN & LO & FS & PL & MA & PT & CO & BX & IT \\
\midrule
CT-CHAT & 0.00 & 0.03 & 0.01 & 0.00 & 0.03 & 0.00 & 0.01 & 0.29 & 0.31 & 0.06 & 0.49 & 0.11 & 0.01 & 0.04 & 0.01 & 0.04 & 0.16 & 0.00 \\
\rowcolor{gray!5}
\quad + MedScribe & 0.04 & 0.29 & 0.17 & 0.22 & 0.39 & 0.00 & 0.01 & 0.28 & 0.33 & 0.46 & 0.45 & 0.33 & 0.46 & 0.18 & 0.10 & 0.28 & 0.16 & 0.16 \\
Gemini & 0.13 & 0.44 & 0.09 & 0.02 & 0.07 & 0.02 & \textbf{0.12} & 0.15 & 0.32 & 0.44 & \textbf{0.60} & 0.31 & 0.47 & 0.00 & 0.06 & \textbf{0.47} & \textbf{0.24} & 0.13 \\
\rowcolor{gray!5}
\quad + MedScribe & 0.15 & 0.48 & 0.23 & 0.28 & 0.45 & 0.02 & 0.08 & 0.22 & 0.31 & 0.44 & 0.48 & 0.32 & 0.54 & 0.13 & \textbf{0.20} & 0.39 & 0.21 & 0.17 \\
MedGemma & \textbf{0.17} & 0.02 & 0.10 & 0.17 & 0.01 & \textbf{0.15} & 0.00 & 0.12 & 0.19 & 0.02 & 0.02 & 0.22 & 0.44 & 0.00 & 0.00 & 0.10 & 0.05 & 0.01 \\
\rowcolor{gray!5}
\quad + MedScribe & 0.15 & 0.56 & \textbf{0.34} & 0.38 & 0.60 & 0.10 & 0.00 & 0.28 & \textbf{0.33} & 0.46 & 0.42 & \textbf{0.33} & 0.74 & 0.19 & 0.11 & 0.30 & 0.07 & 0.12 \\
VILA & 0.04 & 0.18 & 0.00 & 0.00 & 0.00 & 0.01 & 0.01 & 0.00 & 0.01 & 0.04 & 0.20 & 0.00 & 0.01 & 0.00 & 0.00 & 0.00 & 0.00 & 0.13 \\
\rowcolor{gray!5}
\quad + MedScribe & 0.05 & 0.29 & 0.15 & 0.26 & 0.37 & 0.00 & 0.01 & 0.17 & 0.15 & 0.23 & 0.24 & 0.21 & 0.32 & 0.16 & 0.09 & 0.19 & 0.01 & 0.08 \\
MedScribe (ours) & 0.13 & \textbf{0.57} & 0.32 & \textbf{0.44} & \textbf{0.70} & 0.00 & 0.00 & \textbf{0.30} & 0.33 & \textbf{0.46} & 0.48 & 0.30 & \textbf{0.77} & \textbf{0.22} & 0.16 & 0.39 & 0.06 & \textbf{0.18} \\
\midrule
\rowcolor{gray!10}
\multicolumn{19}{c}{\textbf{\textit{RadChestCT}}} \\
\midrule
\rowcolor{gray!10}
Model & MM & AW & CM & PE & CA & HH & LA & EM & AT & LN & LO & FS & PL & MA & PT & CO & BX & IT \\
\midrule
CT-CHAT & 0.00 & 0.05 & 0.00 & 0.00 & 0.05 & 0.00 & 0.00 & 0.15 & 0.39 & 0.42 & \textbf{0.76} & 0.43 & 0.00 & 0.00 & 0.05 & 0.07 & 0.16 & 0.00 \\
\rowcolor{gray!5}
\quad + MedScribe & 0.00 & 0.27 & 0.00 & 0.12 & 0.26 & 0.00 & 0.00 & 0.31 & \textbf{0.48} & 0.55 & 0.71 & \textbf{0.54} & 0.42 & \textbf{0.10} & 0.06 & \textbf{0.27} & 0.09 & 0.12 \\
Gemini & 0.28 & 0.39 & 0.06 & 0.07 & 0.07 & 0.00 & \textbf{0.25} & 0.38 & 0.26 & \textbf{0.65} & 0.71 & 0.49 & 0.55 & 0.00 & 0.00 & 0.25 & \textbf{0.36} & \textbf{0.15} \\
\rowcolor{gray!5}
\quad + MedScribe & 0.27 & 0.38 & 0.11 & 0.27 & 0.43 & 0.00 & 0.18 & 0.38 & 0.37 & 0.51 & 0.63 & 0.40 & 0.63 & 0.10 & \textbf{0.11} & 0.22 & 0.33 & 0.05 \\
MedGemma & \textbf{0.32} & 0.00 & 0.11 & 0.21 & 0.00 & \textbf{0.23} & 0.00 & 0.07 & 0.16 & 0.11 & 0.01 & 0.00 & 0.45 & 0.00 & 0.00 & 0.10 & 0.03 & 0.00 \\
\rowcolor{gray!5}
\quad + MedScribe & 0.28 & 0.32 & \textbf{0.18} & 0.31 & 0.49 & 0.18 & 0.00 & 0.33 & 0.33 & 0.31 & 0.34 & 0.38 & 0.53 & 0.05 & 0.05 & 0.11 & 0.03 & 0.07 \\
VILA & 0.05 & 0.21 & 0.00 & 0.00 & 0.00 & 0.04 & 0.00 & 0.00 & 0.02 & 0.07 & 0.36 & 0.00 & 0.03 & 0.00 & 0.00 & 0.00 & 0.00 & 0.05 \\
\rowcolor{gray!5}
\quad + MedScribe & 0.03 & 0.29 & 0.05 & 0.21 & 0.38 & 0.04 & 0.00 & 0.25 & 0.22 & 0.24 & 0.35 & 0.30 & 0.39 & 0.04 & 0.09 & 0.11 & 0.00 & 0.10 \\
MedScribe (ours) & 0.02 & \textbf{0.47} & 0.16 & \textbf{0.52} & \textbf{0.72} & 0.00 & 0.00 & \textbf{0.42} & 0.44 & 0.39 & 0.63 & 0.40 & \textbf{0.74} & 0.05 & 0.09 & 0.26 & 0.08 & 0.14 \\
\midrule
\end{tabular}}
\small \textit{Pathology abbreviations: MM: Medical material, AW: Arterial wall calcification, CM: Cardiomegaly, PE: Pericardial effusion, CA: Coronary artery wall calcification, HH: Hiatal hernia, LA: Lymphadenopathy, EM: Emphysema, AT: Atelectasis, LN: Lung nodule, LO: Lung opacity, FS: Pulmonary fibrotic sequela, PL: Pleural effusion, MA: Mosaic attenuation pattern, PT: Peribronchial thickening, CO: Consolidation, BX: Bronchiectasis, IT: Interlobular septal thickening.}
\label{tab:per_pathology_f1}

\end{table*}

\begin{table*}[t]
\centering
\caption{Summary metrics: CT-RATE (classification and text generation).}
\label{tab:summary_metrics_ctrate}
\resizebox{0.8\textwidth}{!}{%
\small
\begin{tabular}{l c c c c c c}
\midrule
\rowcolor{gray!10}
\multicolumn{7}{c}{CT-RATE} \\
\midrule
\rowcolor{gray!10}
Model & Macro P & Macro R & Macro F1 & BLEU-1 & METEOR & ROUGE-L \\
\midrule
CT-CHAT & 0.27 & 0.11 & 0.09 & 0.30 & \textbf{0.34} & \textbf{0.29} \\
\rowcolor{gray!5}
\quad + MedScribe & 0.34 & 0.23 & 0.24 & 0.30 & 0.33 & 0.27 \\
Gemini & 0.33 & 0.25 & 0.23 & 0.19 & 0.21 & 0.12 \\
\rowcolor{gray!5}
\quad + MedScribe & 0.36 & 0.27 & 0.28 & 0.19 & 0.20 & 0.12 \\
MedGemma & \textbf{0.42} & 0.07 & 0.10 & 0.02 & 0.11 & 0.07 \\
\rowcolor{gray!5}
\quad + MedScribe & 0.34 & 0.31 & 0.31 & 0.10 & 0.17 & 0.12 \\
VILA & 0.21 & 0.03 & 0.04 & 0.03 & 0.11 & 0.08 \\
\rowcolor{gray!5}
\quad + MedScribe & 0.36 & 0.12 & 0.17 & 0.00 & 0.06 & 0.05 \\
MedScribe (ours) & 0.36 & \textbf{0.33} & \textbf{0.32} & \textbf{0.34} & 0.34 & 0.18 \\
\midrule
\end{tabular}}
\end{table*}
\begin{table*}[t]
\centering
\caption{Summary metrics: RadChestCT (classification only).}
\label{tab:summary_metrics_radchest}
\small
\begin{tabular}{l c c c}
\midrule
\rowcolor{gray!10}
\multicolumn{4}{c}{RadChestCT} \\
\midrule
\rowcolor{gray!10}
Model & Macro P & Macro R & Macro F1 \\
\midrule
CT-CHAT & 0.22 & 0.13 & 0.14 \\
\rowcolor{gray!5}
\quad + MedScribe & 0.30 & 0.23 & 0.24 \\
Gemini & \textbf{0.44} & 0.27 & 0.27 \\
\rowcolor{gray!5}
\quad + MedScribe & 0.38 & 0.29 & 0.30 \\
MedGemma & 0.29 & 0.10 & 0.10 \\
\rowcolor{gray!5}
\quad + MedScribe & 0.33 & 0.23 & 0.24 \\
VILA & 0.25 & 0.04 & 0.05 \\
\rowcolor{gray!5}
\quad + MedScribe & 0.38 & 0.13 & 0.17 \\
MedScribe (ours) & 0.39 & \textbf{0.32} & \textbf{0.31} \\
\midrule
\end{tabular}
\end{table*}

\subsection{Baseline slice-count ablation and evaluation protocol}
\label{supp:baselines}

For the 2D-based baselines MedGemma and Gemini, we conducted an ablation over the number of axial slices provided as input. Table~\ref{tab:supp_slice_ablation} reports the macro F1 and BLEU-1 scores on CT-RATE across the tested configurations. For both models, classification performance improves with an increasing number of slices, with the best macro F1 reached at $n_{\text{slices}} = 128$. We did not extend the ablation to $n_{\text{slices}} = 256$: the MedGemma 1.5 technical report \cite{sellergren2026medgemma15technicalreport} indicates that the model was trained and evaluated with a maximum of 85 axial slices per query, and $n_{\text{slices}} = 128$ already exceeds this range. The best configuration ($n_{\text{slices}} = 128$) was adopted for both models to keep the comparison fair.

\medskip
\noindent\textbf{Note on evaluation protocol.} MedGemma's own CT-RATE evaluation in \cite{sellergren2026medgemma15technicalreport} is conducted as a per-condition binary classification task, where the model is independently queried for the presence or absence of each pathology and evaluated on the resulting binary answer. In contrast, our evaluation protocol targets full radiology report generation: each model produces a free-text report in a single pass, from which pathology labels are subsequently derived via a RadBERT-based classifier. These are strictly different tasks, and MedGemma's macro F1 values in our setting are expected to be lower than those reported in \cite{sellergren2026medgemma15technicalreport}. Our choice reflects the end-to-end clinical reporting scenario targeted by MedScribe, rather than a set of isolated diagnostic queries.

\begin{table}[h]
\caption{Slice-count ablation for the 2D-based baselines MedGemma 1.5-4B-it and Gemini 2.5 Flash on CT-RATE. Macro F1 is reported across the 18 pathologies. Bold indicates the best configuration per model.}
\centering
\begin{tabular}{l c c c}
\midrule
\rowcolor{gray!10}
Model & $n_{\text{slices}}$ & Macro F1 & BLEU-1 \\
\midrule
\multirow{4}{*}{MedGemma 1.5-4B-it} 
 & 16  & 0.075 & 0.053 \\
 & 32  & 0.090 & \textbf{0.086} \\
 & 64  & 0.084 & 0.078 \\
 & 128 & \textbf{0.099} & 0.020 \\
\midrule
\multirow{4}{*}{Gemini 2.5 Flash} 
 & 16  & 0.225 & 0.190 \\
 & 32  & 0.221 & \textbf{0.192} \\
 & 64  & 0.240 & 0.194 \\
 & 128 & \textbf{0.258} & 0.190 \\
\midrule
\end{tabular}
\label{tab:supp_slice_ablation}
\end{table}

\subsection{Sensitivity analysis on the number of retrieved neighbors $k$}
\label{supp:k_ablation}
The number of neighbors $k$ retrieved in the RAG space controls the volume of context provided to the LLM. To validate the default $k=3$, we evaluated the MedScribe-Template configuration across $k \in \{1, 3, 5, 7, 9, 11, 13\}$ on a random subset of 256 CT-RATE validation cases, holding all other pipeline components constant. Table~\ref{tab:supp_k_ablation} reports the macro F1 and text-generation metrics.

Performance remains stable across the tested range. Macro F1 peaks at $k=3$ (0.355), with a maximum deviation of 0.012 across all configurations. While text metrics (BLEU-1, ROUGE-L, METEOR) show a slight decreasing trend as $k$ increases, they remain competitive. We retained $k=3$ because it optimizes macro F1 while minimizing the number of tokens per inference, thereby reducing computational latency.
 
\begin{table}[h]
\caption{Sensitivity analysis on the number of retrieved neighbors $k$ in the multidimensional RAG space. Macro F1 and text-generation metrics are reported for MedScribe-Template on a fixed random subset of 256 CT-RATE validation cases. Bold indicates the best value per column; the chosen default $k=3$ is underlined. Absolute scores may differ from those reported in the main paper because of the reduced evaluation subset; relative trends across $k$ are the quantity of interest here.}
\centering
\begin{tabular}{c c ccc}
\midrule
\rowcolor{gray!10}
$k$ & Macro F1 & BLEU-1  & ROUGE-L & METEOR \\
\midrule
1              & 0.354          & \textbf{0.355} & \textbf{0.190} & \textbf{0.343} \\
\underline{3}  & \textbf{0.355} & 0.330                   & 0.185          & 0.330          \\
5              & 0.354          & 0.327                   & 0.184          & 0.329          \\
7              & 0.351          & 0.324                   & 0.184          & 0.330          \\
9              & 0.343          & 0.322                   & 0.184          & 0.330          \\
11             & 0.355          & 0.327                   & 0.185          & 0.328          \\
13             & 0.350          & 0.322                   & 0.183          & 0.328          \\
\midrule
\end{tabular}
\label{tab:supp_k_ablation}
\end{table}

\subsection{Snippet extraction prompt and few-shot examples}
\label{supp:prompt}
To construct the textual component of the multidimensional RAG reference space, we extract pathology-specific sentences from each CT-RATE training report using a general-purpose LLM (Llama 3.1-8B) with four-shot prompting. For each of the 18 target pathologies, we designed a dedicated prompt consisting of (i) a fixed instruction template, shared across all pathologies and parameterized by the pathology name, (ii) a user level message wrapping four human curated examples and the target report, and (iii) a set of four examples randomly selected from the CT-RATE training set to cover both positive and negative findings. All prompts and example sets will be released together with the source code for full reproducibility.

The extraction procedure follows a strict output policy: when the target pathology is not mentioned in the report, the model is instructed to return a canonical no finding sentence (``No sign of \{pathology\} was found in the scan.''), enabling unambiguous downstream parsing of negative cases. When multiple pathologies are discussed in the same sentence, only the clause directly describing the target pathology is extracted, preserving sentence-level pathology specificity. Figure \ref{fig:prompt-template} and Figure \ref{fig:prompt-examples} show the full system instruction template, the user-message wrapper, and the four few-shot examples for one representative pathology (arterial wall calcification); the examples for the remaining pathologies follow the same structure and are available in the released repository.

\begin{figure}
\begin{tcolorbox}[
    colback=gray!5,
    colframe=gray!60,
    title=\textbf{System Instruction Template},
    fonttitle=\bfseries,
    breakable,
    enhanced
]
\small\ttfamily
You are a precise radiology report information extractor specialized in identifying information about \{pathology\}. \\[0.3em]
Your goal: \\[0.3em]
Extract only the sentences from a chest CT report that describe \{pathology\}. \\[0.3em]
Rules: \\
- Include both positive findings (presence of \{pathology\}) and negative findings (absence of \{pathology\}). \\
- Extract sentences only from the Findings section of the report. \\
- Do not include unrelated findings. \\
- Return exact sentences from the report — do not paraphrase, summarize, or infer. \\
- If a sentence discusses multiple pathologies, extract only the content related to \{pathology\}. \\
- If multiple sentences discuss \{pathology\}, extract all of them. \\
- If information about \{pathology\} is not present, output exactly: \\
\hspace*{1em}\textit{No sign of \{pathology\} was found in the scan.} \\
\end{tcolorbox}

\begin{tcolorbox}[
    colback=gray!5,
    colframe=gray!60,
    title=\textbf{User Message Template},
    fonttitle=\bfseries,
    breakable,
    enhanced
]
\small\ttfamily
{[EXAMPLES]} \\
\{examples\} \\
\\
{[TASK]} \\
Extract information from the following chest CT report. \\[0.3em]
Report: \\
\{report\} \\
\\
Extraction:
\end{tcolorbox}
\caption{System instruction template and user message wrapper used for snippet extraction, shared across all 18 pathologies.}
\label{fig:prompt-template}
\end{figure}

\begin{figure}[htbp]
\begin{tcolorbox}[
    colback=blue!3,
    colframe=blue!40,
    title=\textbf{Few-Shot Examples (Arterial Wall Calcification)},
    fonttitle=\bfseries,
    breakable,
    enhanced
]
\small\ttfamily
\textbf{Example 1:} \{\\
\hspace*{1em}"report": "Findings: No lymph node was observed in the supraclavicular fossa ... There are calcified atheroma plaques in the thoracic and abdominal aorta ...",\\
\hspace*{1em}"extraction": "There are calcified atheroma plaques in the thoracic and abdominal aorta."\\
\} \\[0.5em]

\textbf{Example 2:} \{\\
\hspace*{1em}"report": "Findings: Trachea, both main bronchi are open. Calcific atheroma plaques are observed in the aorta and coronary arteries ...",\\
\hspace*{1em}"extraction": "Calcific atheroma plaques are observed in the aorta."\\
\} \\[0.5em]

\textbf{Example 3:} \{\\
\hspace*{1em}"report": "Findings: There is a centrally located primary tumoral lesion in the upper lobe of the left lung ...",\\
\hspace*{1em}"extraction": "No sign of Arterial wall calcification was found in the scan."\\
\} \\[0.5em]

\textbf{Example 4:} \{\\
\hspace*{1em}"report": "Findings: Trachea and both main bronchi are open. There are peripheral and centrally located ground-glass areas ...",\\
\hspace*{1em}"extraction": "No sign of Arterial wall calcification was found in the scan."\\
\}
\end{tcolorbox}
\caption{Four-shot examples for arterial wall calcification. Examples for the remaining pathologies follow the same structure and will be available in the released repository.}
\label{fig:prompt-examples}
\end{figure}

\medskip
\noindent\textbf{Empirical verification of extraction quality.}
To provide an empirical check on the reliability of the LLM-based snippet extraction, we validated the extracted pathology-specific snippets against the CT-RATE ground truth labels using two independent verification methods. The first, which we denote \textit{Classifier F1}, consists of running the same RadBERT-based classifier used in the main evaluation on each extracted snippet and comparing its predicted label with the CT-RATE ground-truth label. The second, denoted \textit{Template F1}, is a rule-based check that labels a snippet as ``absent'' if it matches the canonical negative-finding sentence prescribed by the prompt (``No sign of \{pathology\} was found in the scan.''), and as ``present'' otherwise.

The two methods are complementary and fail in different, identifiable ways. The Classifier F1 is sensitive to lexical shortcuts: when a snippet contains the pathology name in a negative context (e.g., ``No sign of emphysema was found in the scan.''), the classifier occasionally fires positive due to pathology name recognition, yielding systematically low precision on pathologies with weaker training signal. The Template F1, conversely, is sensitive to non-canonical negation phrasing: when the LLM extracts a clinically negative sentence that does not match the prescribed template (e.g., ``Heart size is normal.'' for cardiomegaly, or ``No pleural effusion was observed.'' for pleural effusion), the rule treats it as ``present'' even though the clinical meaning is negative. These two failure modes are structurally distinct and unlikely to co-occur on the same snippet, so taking the per-pathology maximum of the two F1 scores provides a principled lower bound on the true extraction quality. Table~\ref{tab:supp_extraction} reports both metrics alongside their maximum for each of the 17 target pathologies.

Under this criterion, extraction quality is high across the full pathology set: the per-pathology maximum F1 exceeds 0.90 on 14 of 17 pathologies, with a minimum of 0.74 (bronchiectasis) and a mean of 0.906. This confirms that the LLM-based extraction is reliable across the full pathology set, and that the residual noise introduced by the extraction step is small compared to the downstream variability of the report generation task.

\begin{table}[h]
\caption{Empirical verification of snippet extraction quality on the CT-RATE training split. Classifier F1 uses a RadBERT-based classifier applied to each extracted snippet; Template F1 uses a rule-based match against the canonical negative-finding sentence. The two methods fail in complementary ways (see text); the per-pathology maximum of the two scores provides a lower bound on the true extraction quality.}
\centering
\begin{tabular}{l c c c}
\midrule
\rowcolor{gray!10}
Pathology & Classifier F1 & Template F1 & Max F1 \\
\midrule
Arterial wall calcification           & 0.928 & 0.920 & 0.928 \\
Atelectasis                           & 0.981 & 0.908 & 0.981 \\
Bronchiectasis                        & 0.189 & 0.741 & 0.741 \\
Cardiomegaly                          & 0.954 & 0.325 & 0.954 \\
Consolidation                         & 0.958 & 0.694 & 0.958 \\
Coronary artery wall calcification    & 0.977 & 0.922 & 0.977 \\
Emphysema                             & 0.343 & 0.832 & 0.832 \\
Hiatal hernia                         & 0.251 & 0.982 & 0.982 \\
Interlobular septal thickening        & 0.148 & 0.919 & 0.919 \\
Lung nodule                           & 0.955 & 0.870 & 0.955 \\
Lung opacity                          & 0.907 & 0.694 & 0.907 \\
Medical material                      & 0.916 & 0.861 & 0.916 \\
Mosaic attenuation pattern            & 0.147 & 0.777 & 0.777 \\
Peribronchial thickening              & 0.191 & 0.794 & 0.794 \\
Pericardial effusion                  & 0.935 & 0.165 & 0.935 \\
Pleural effusion                      & 0.917 & 0.299 & 0.917 \\
Pulmonary fibrotic sequela            & 0.422 & 0.926 & 0.926 \\
\midrule
\textbf{Mean}                          & \textbf{0.666} & \textbf{0.743} & \textbf{0.906} \\
\midrule
\end{tabular}
\label{tab:supp_extraction}
\end{table}

\subsection{Statistical analysis of per-pathology improvements}
\label{supp:stat_analysis}

To assess whether the per-pathology F1 improvements of MedScribe over the general-purpose VLM baselines (CT-CHAT, Gemini 2.5 Flash, MedGemma, and VILA-M3) are statistically significant, we fit a linear mixed-effects model on the per-pathology F1 differences

\[
\Delta_{i,m} = F1^{\text{MedScribe}}_{i} - F1^{\text{baseline}_m}_{i},
\]

where $i$ indexes the 18 CT-RATE pathologies and $m$ indexes the four baseline VLMs, yielding $18 \times 4 = 72$ observations per analysis. The model contains a per-pathology fixed-effect intercept and a random intercept on the baseline identity $m$:

\[
\Delta_{i,m} = \alpha_i + u_m + \varepsilon_{i,m}, \qquad u_m \sim \mathcal{N}(0,\sigma_u^2), \quad \varepsilon_{i,m} \sim \mathcal{N}(0,\sigma_\varepsilon^2).
\]

The fixed effect $\alpha_i$ estimates the mean improvement of MedScribe over the VLM baselines for pathology $i$, while the random effect $u_m$ captures systematic differences in the baseline strength across VLMs (e.g., CT-CHAT being weaker on average than Gemini), so that significance is not inflated by between-baseline correlation. We fit the model via restricted maximum likelihood (REML) using \texttt{statsmodels} \cite{seabold2010statsmodels}, and test the per-pathology null hypothesis $H_0: \alpha_i = 0$ via the corresponding Wald $z$-statistic. The procedure is applied separately for the MedScribe-Template and MedScribe-VLM configurations; per-pathology improvements with $p < 0.05$ are flagged as statistically significant in Figure~\ref{fig:performance}.

The 95\% confidence intervals from the model are uniform across pathologies within each configuration (half-width $\approx 0.11$ for MedScribe-VLM and $\approx 0.14$ for MedScribe-Template), and exclude zero for the pathologies flagged as statistically significant in Figure~\ref{fig:performance}.

\subsection{Inference cost and latency analysis}
\label{supp:cost}

To assess the practical cost of deploying MedScribe relative to general-purpose VLM baselines, we measured per-scan latency and per-scan token consumption on a fixed sample of CT-RATE scans. All locally hosted models were run on a single NVIDIA H100 GPU; Gemini 2.5 Flash was queried via the public API, so its reported latency includes network round-trip time and service-side queuing. Table~\ref{tab:supp_cost} reports the resulting averages.

A key observation is that the total token budget is not directly comparable across systems, because it is composed of fundamentally different elements. For MedScribe, the LLM consumes text tokens only -- quantitative descriptors produced by the diagnostic tools and snippets retrieved from the multidimensional RAG space -- with no visual tokens consumed at any stage of the agent loop. For 2D-based VLMs (Gemini 2.5 Flash, MedGemma 1.5-4B-it, VILA-M3), the per-scan token budget is instead dominated by image tokens: with $n_{\text{slices}} = 128$ axial slices and approximately 256 vision tokens per slice, more than 30{,}000 image tokens are consumed for each scan, in addition to the textual prompt. CT-CHAT follows a different architecture in which the visual signal is encoded into a fixed-size embedding by CT-CLIP before reaching the language decoder, so its visual cost is structurally distinct from the per-slice tokenization of the 2D VLMs. The numerical comparison should therefore be interpreted in light of these architectural differences.

The most relevant takeaway is that tool-grounded reasoning decouples per-scan inference cost from raw imaging input size: MedScribe's token consumption scales with retrieval and reasoning complexity rather than with the number of slices or image resolution, which is a desirable property for high-resolution or long-acquisition volumetric studies. The wall-clock latency of MedScribe is dominated by the segmentation step (TotalSegmentator) followed by the agent loop; the LLM-only inference time is comparable to that of MedGemma, despite the agent issuing an average of 14 tool calls per scan with the associated context-accumulation overhead.

\begin{table}[h]
\caption{Inference cost and latency comparison on CT-RATE. Latency is reported as average wall-clock time per scan; tokens per scan are total tokens consumed by the model per generated report (input + output, including image tokens for 2D-VLM baselines). $^*$Gemini 2.5 Flash is queried via the public API, so its latency includes network round-trip time. $^\dagger$MedScribe latency excludes segmentation time, which adds approximately [SEG\_TIME]\,s per scan when running TotalSegmentator on the same hardware. CT-CHAT consumes its visual input through a fixed-size CT-CLIP embedding (no per-slice tokenization), so its image-token count is not directly comparable to the 2D VLMs.}
\centering
\begin{tabular}{l c c c}
\midrule
\rowcolor{gray!10}
Model & Hardware & Avg. time per scan (s) & Avg. tokens per scan \\
\midrule
MedScribe-Template$^\dagger$ (ours) & H100 & 17.57 + TS & 11{,}527 \\
MedGemma 1.5-4B-it                  & H100 & 19.36 & 33{,}776 \\
CT-CHAT                             & H100 & 7.62 & 14{,}075 \\
Gemini 2.5 Flash$^*$                & API  & 2.13 & 33{,}576
 \\
\midrule
\end{tabular}
\label{tab:supp_cost}
\end{table}

\subsection{Sensitivity to the LLM backbone}
\label{supp:llm_backbone}

The choice of Llama 3.1-8B as the LLM backbone for all three roles in MedScribe was made deliberately to match the LLM used by CT-CHAT \cite{hamamci2024developing}, our most direct 3D-native baseline. This design choice isolates the contribution of the agentic, tool-grounded framework from any improvement that might be attributable purely to a stronger underlying LLM, and supports the claim that MedScribe's gains over CT-CHAT come from the framework rather than from a backbone advantage.

To verify that the framework's behaviour generalizes beyond this specific scale and to address the natural question of whether a larger LLM would change the conclusions, we additionally tested MedScribe-Template on CT-RATE with Llama 3.3-70B as the LLM backbone, keeping every other component (tools, RAG retrieval space, prompts, $k=3$) unchanged. Table~\ref{tab:supp_llm_backbone} compares the two configurations.

The 70B backbone yields a meaningful improvement in classification (macro F1 increases from 0.323 to 0.353, $+9\%$ relative), driven by uniformly higher precision and recall, while text-generation metrics remain essentially unchanged (BLEU-1, ROUGE-L, and METEOR all within $0.002$ of the 8B values). This pattern indicates that linguistic fluency in MedScribe is largely saturated already at the 8B scale, while reasoning over the tool outputs and retrieved neighbors continues to benefit from a stronger LLM. The cost is a $\approx 2.7\times$ increase in per-scan latency (from 17.57s to 48.15s on the same H100 GPU), which is a non-trivial tradeoff in deployment scenarios. Importantly, the qualitative conclusions of the paper -- that MedScribe outperforms general-purpose VLMs and approaches task-specific fine-tuning without any task-specific training -- are preserved and strengthened with the larger backbone, confirming that the framework benefits monotonically from stronger LLMs rather than relying on a particular backbone choice.

\begin{table}[h]
\caption{Effect of the LLM backbone on MedScribe-Template performance on CT-RATE. Every component other than the LLM is kept unchanged. Classification metrics are macro-averaged across the 18 CT-RATE pathologies.}
\centering
\begin{tabular}{l c c c c c c}
\midrule
\rowcolor{gray!10}
LLM backbone & Macro F1 & Macro P. & Macro R. & BLEU-1 & ROUGE-L & METEOR \\
\midrule
Llama 3.1-8B (default) & 0.323 & 0.363 & 0.328 & 0.341 & 0.183 & \textbf{0.336} \\
Llama 3.3-70B          & \textbf{0.353} & \textbf{0.398} & \textbf{0.394} & \textbf{0.342} & \textbf{0.185} & 0.334 \\
\midrule
\multicolumn{7}{l}{\textit{Inference latency on H100:} 8B = 17.57 s/scan; 70B = 48.15 s/scan ($\approx 2.7\times$ slower).} \\
\midrule
\end{tabular}
\label{tab:supp_llm_backbone}
\end{table}

\bibliographystyle{unsrtnat}
\bibliography{references}  






\end{document}